\documentclass{article}

\usepackage{arxiv}

\usepackage[utf8]{inputenc} 
\usepackage[T1]{fontenc}    
\usepackage{hyperref}       
\usepackage{url}            
\usepackage{booktabs}       
\usepackage{amsfonts}       
\usepackage{nicefrac}       
\usepackage{microtype}      
\usepackage{lipsum}		
\usepackage{graphicx}
\usepackage{doi}
\usepackage{multirow}
\usepackage{amsmath}
\usepackage[normalem]{ulem}
\usepackage[backend=biber, style=numeric, maxnames=50, minnames=10]{biblatex}
\usepackage{fontawesome}
\usepackage{xcolor} 

\newcommand{\modelnamepreview}{Megrez2-Preview}
\newcommand{\modelname}{Megrez2}

\newcommand{\dotitem}[1]{%
  \par
  \begingroup
  \parindent=0pt
  \leftskip=-0.5em 
  \parshape 2 2em \dimexpr\linewidth-2em\relax 2em \dimexpr\linewidth-2em\relax
  \makebox[1em][l]{\raisebox{0.2ex}{$\bullet$}}%
  #1\par
  \endgroup
}

\title{Megrez2 \ Technical Report}

\author{\bf
    Boxun Li$^1$ \quad Yadong Li$^1$ \quad Zhiyuan Li$^1$ \quad Congyi Liu$^1$ \quad Weilin Liu$^1$ \quad Guowei Niu$^1$ \quad \\
    \bf
    Zheyue Tan$^{1,2}$ \quad Haiyang Xu$^1$ \quad Zhuyu Yao$^1$ \quad Tao Yuan$^1$ \quad Dong Zhou$^1$ \quad Yueqing Zhuang$^1$ \quad \\
    \bf
    Bo Zhao$^2$ \quad Guohao Dai$^{3,}$\thanks{Corresponding authors.} \quad Yu Wang$^{4,\ast}$ \\
    \normalsize \textsuperscript{1} Infinigence-AI\thanks{The listing of authors is in alphabetical order based on their last names.} \quad
    \normalsize \textsuperscript{2} Aalto University \quad
    \normalsize \textsuperscript{3} Shanghai Jiao Tong University \quad
    \normalsize \textsuperscript{4} Tsinghua University
}

\date{}

\renewcommand{\headeright}{}
\renewcommand{\undertitle}{}

\begin{document}
\maketitle

\begin{center}
  \vspace{-3em}
  \faGithub~\url{https://github.com/infinigence/Infini-Megrez}
  \vspace{0.75em}
\end{center}

\begin{abstract}

We present \modelname, a novel lightweight and high-performance language model architecture optimized for device native deployment. 
\modelname~introduces a novel cross-layer expert sharing mechanism, which significantly reduces total parameter count by reusing expert modules across adjacent transformer layers while maintaining most of the model's capacity. 
It also incorporates pre-gated routing, enabling memory-efficient expert loading and faster inference.
As the first instantiation of the \modelname~architecture, we introduce the \modelnamepreview~model, which is pre-trained on a 5-trillion-token corpus and further enhanced through supervised fine-tuning and reinforcement learning with verifiable rewards.
With only 3B activated and 7.5B stored parameters, \modelnamepreview~demonstrates competitive or superior performance compared to larger models on a wide range of tasks, including language understanding, instruction following, mathematical reasoning, and code generation.
These results highlight the effectiveness of the \modelname~architecture to achieve a balance between accuracy, efficiency, and deployability, making it a strong candidate for real-world, resource-constrained applications.

\end{abstract}

\vspace{-1em}
\begin{figure}[!h]
    \centering
    \includegraphics[width=0.7\linewidth]{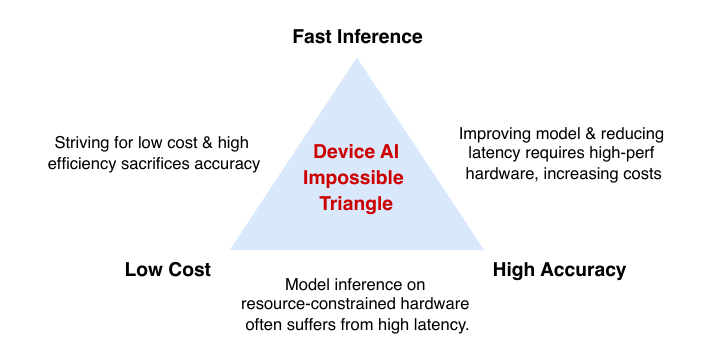}
    \caption{The Impossible Triangle of Device AI}
    \label{fig:triangle}
\end{figure}

\section{Introduction}
Large Language Models (LLMs) have progressed at an unprecedented pace, scaling to hundreds of billions of parameters and demonstrating remarkable advances toward Artificial General Intelligence (AGI).
Recent foundation models such as GPT-4~\cite{achiam2023gpt}, GPT-o3~\cite{openai_gpto3_o4}, Gemini~\cite{geminiteam2025geminifamilyhighlycapable}, Llama 4~\cite{meta2025llama4}, DeepSeekR1~\cite{guo2025deepseek}, Kimi-K2~\cite{kimi-k2}, and the Qwen3 series~\cite{Yang2025Qwen3TR} have exhibited strong performance across complex tasks in multiple domains.
This progress has been driven by massive investments in data and compute, but such growth also intensifies the tension between model capacity and deployment practicality—particularly on devices with strict latency, memory, and power constraints.

Given the substantial costs involved, Mixture-of-Experts (MoE) architectures have become an increasingly attractive alternative.
By dynamically activating only a subset of specialized experts per input, MoEs offer a compelling trade-off: they can match or even exceed the performance of dense models while significantly reducing inference-time computational demands.
However, conventional MoE designs face notable deployment challenges on resource-constrained platforms.
Sparse expert activations can lead to high memory usage and fragmented parameter utilization, thereby limiting the ability of on-device hardware to serve such models efficiently.
Moreover, the design of device AI systems often encounters the \textit{Impossible Triangle}: a three-way trade-off among speed, accuracy, and cost, where enhancing one aspect typically compromises the others. 
\autoref{fig:triangle} illustrates these inherent trade-offs in deploying models on device native hardware. 
Accelerating inference on constrained hardware usually demands aggressive pruning or quantization, which reduces model capacity and thereby degrades accuracy. 
On the other hand, increasing model size to boost accuracy leads to higher latency and greater hardware demands, which can be problematic for battery-powered or thermally constrained devices. 
This \textit{device-model trilemma} delineates a design frontier where progress along one dimension often comes at the expense of the other two.

To address these challenges, we propose \modelname, a novel language model architecture specifically designed for device native deployment.
The core innovation of our approach is a \textbf{cross-layer expert sharing mechanism}: by reusing the same set of experts across multiple adjacent layers, \modelname~significantly reduces the total parameter count while maintaining the number of activated parameters—crucial for preserving model performance.
Combined with \textbf{pre-gated routing} and a novel routing strategy that encourages balanced expert utilization, our method achieves strong performance under tight computational constraints.
This cross-layer sharing not only enhances parameter efficiency but also improves hardware utilization, making \modelname~particularly well-suited for resource-constrained devices. 

In the following sections, we first review related work. We then describe the \modelname~architecture. To demonstrate the effectiveness of our approach, we present the training methodology and evaluation results of \modelnamepreview, which serves as the first instantiation of the \modelname~architecture. Finally, we conclude with a summary of our work.

\vspace{-1em}
\section{Related Work}

Mixture-of-Experts (MoE) models have emerged as a promising architecture for scaling large language models (LLMs) by decoupling the total parameter count from the computational cost per token. A key trend is the adoption of increasingly granular expert pools, often with shared experts across layers, enabling higher capacity and improved specialization while maintaining efficiency.

DeepSeekMoE~\cite{dai2024deepseekmoe} introduces a fine-grained MoE structure, segmenting each Feed-Forward-Network (FFN) into smaller experts. This enables a larger expert pool without increasing the total parameter count. By pioneering a system-level optimization stack tailored for MoEs, DeepSeekMoE achieves inference speeds up to 4.5 times faster and 9 times cheaper than comparable dense models. Additional optimization techniques such as auxiliary-loss-free routing~\cite{wang2024auxiliary} further enhance the model's routing efficiency.
The Qwen series~\cite{Yang2024Qwen25TR,Yang2025Qwen3TR} explores MoE architectures across various scales. For instance, Qwen1.5-MoE-A2.7B matches the performance of 7B dense models while activating only 2.7B parameters, combining 4 shared and 60 dynamic experts per MoE layer. Expanding on this, Qwen3 introduces models such as the 235B-A22B, providing 235B total parameters with only 22B active per token. These models illustrate how fine-grained MoE coupled with expert sharing significantly reduces computational cost without sacrificing accuracy.
Skywork-MoE~\cite{wei2024skywork}, initialized from a 13B dense model, employs a 146B MoE structure with 22B active parameters per token. Techniques like Gating Logit Normalization (GLN) and adaptive auxiliary losses enhance expert load balancing and specialization, positioning Skywork-MoE competitively alongside large-scale MoEs such as DBRX and Mixtral.

Deploying large language models (LLMs) on devices presents significant challenges due to limited memory and computational resources, making it impractical to run large-scale models directly. In addition to the Qwen series~\cite{Yang2025Qwen3TR}, smaller models such as Gemma~\cite{Kamath2025Gemma3T} and Phi-4-Mini~\cite{abouelenin2025phi} offer viable alternatives tailored for memory-constrained environments. For instance, Phi-4-Mini, with its 3.8 billion parameters, demonstrates reasoning capabilities comparable to—or even surpassing—models with over 7 billion parameters. Google’s Gemma 3~\cite{Kamath2025Gemma3T}, which ranges from 1B to 27B parameters, supports quantized inference (e.g., INT4), making it suitable for single-GPU setups and mobile devices. Notably, the smaller variants, Gemma3-1B and Gemma3-4B, deliver performance on par with the previous generation Gemma2 across multiple benchmarks, illustrating that compact LLMs can offer large-model quality within an device-friendly footprint. Additionally, Megrez-3B-Instruct and Megrez-3B-Omni~\cite{li2025megrez} provide strong small-scale dense models optimized for device native deployment.

Specialized optimizations are critical for deploying MoEs on devices. While MoEs offer improved performance with reduced computation, their overall memory footprint remains significant. Earlier approaches employed offloading techniques to facilitate serving MoE models on memory-constrained devices~\cite{eliseev2023fast}, but such methods remain constrained by memory bandwidth.
Pre-gated MoE~\cite{hwang2024pre} addresses this challenge by introducing a pre-gating function that reduces the dynamic nature of sparse expert activation, thus managing MoE's large memory footprint more effectively. Similarly, Read-ME~\cite{cai2024textit} applies pre-gating strategies in its router, enabling expert-aware batching and lookahead scheduling based on expert workload dispatching, significantly reducing memory usage during MoE inference, especially beneficial for device native deployments.

\section{Model Architecture}
\begin{figure}[t]
    \centering
    \includegraphics[width=1.0\linewidth]{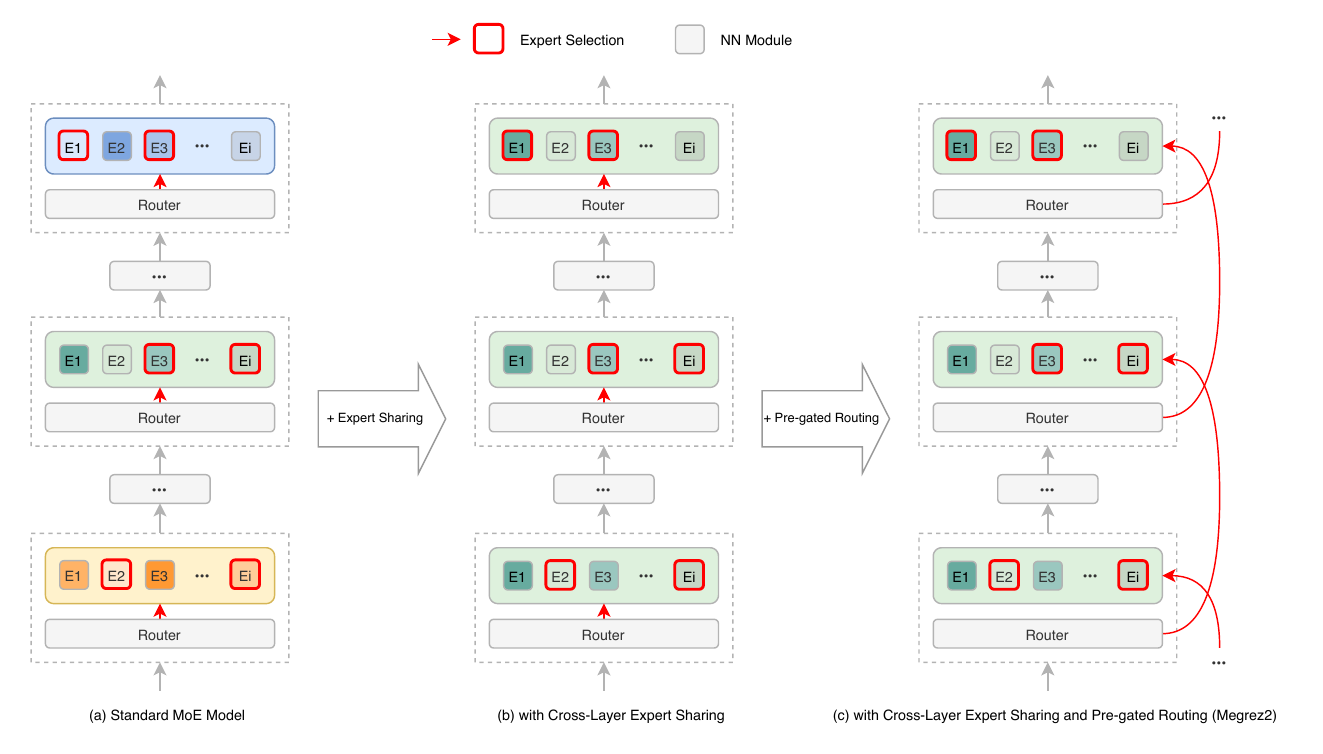}
    \caption{Expert Sharing and Pre-gated Routing in \modelname~Model}
    \label{fig:moe_arch}
\end{figure}

In this section, we present the architecture of \modelname, which incorporates a novel cross-layer expert sharing strategy and pre-gated routing within the Mixture-of-Experts (MoE) framework.

\subsection{Mixture-of-Experts Block}

We first briefly review the widely adopted $\operatorname{top}\text{-}k$ MoE structure~\cite{Yang2025Qwen3TR,dai2024deepseekmoe,liu2024deepseek}.
At layer~$i$, a private gating network~$G_i$ selects, from its dedicated pool of $M$ experts $\{E_{i,1},\dots,E_{i,M}\}$, the $k$ most relevant experts ($k \ll M$).
Given the layer input $\mathbf{h}_i$, the gating network produces routing scores $\mathbf{s}_i = G_i(\mathbf{h}_i)$.
The aggregated hidden state $h_{i+1}$ is then computed as

\begin{equation}
  \mathbf{h}_i' \;=\;
  \sum_{j=1}^{M}
  \bigl[\operatorname{top}\text{-}k(\mathbf{s}_i)\bigr]_j\,
  E_{i,j}\!\bigl(\mathbf{h}_i\bigr),
  \label{eq:std_moe}
\end{equation}

where $\operatorname{top}\text{-}k(\cdot)$ retains only the $k$ largest entries of $\mathbf{s}_i$, setting the others to zero.
Each expert, implemented as a feed-forward network, computes $E_{i,j}(\mathbf{h}_i)$; these outputs are then combined with their corresponding routing weights to produce the block’s output. For clarity, we omit the \emph{shared experts} introduced in DeepSeekMoE~\cite{dai2024deepseekmoe}, which are always active in every layer.
As shown in \autoref{fig:moe_arch}(a), the router in each layer computes the gating scores independently and selects the corresponding experts to produce the output of this block.

\subsection{Cross-Layer Expert Sharing}

To reduce the parameter count of a standard MoE, we share each expert’s parameters across $n$ consecutive layers, which lowers the total parameters by roughly a factor of~$n$. 
Specifically, \modelname~partitions the $L$-layer transformer into $G = L/n$ contiguous \emph{groups} of length~$n$. 
Every layer $i$ in group $g = \lfloor i/n \rfloor$ \emph{shares} the same expert pool ${E_{g,1},\dots,E_{g,M}}$ while retaining its own gating network $G_i$ and projection weights. Layer $i$ then updates its hidden state as

\begin{equation}
  \mathbf{h}_i' \;=\;
  \sum_{j=1}^{M}
  \bigl[\operatorname{top}\text{-}k(\mathbf{s}_i)\bigr]_j\,
  E_{g,j}\!\bigl(\mathbf{h}_i\bigr),
  \qquad
  g = \bigl\lfloor i / n \bigr\rfloor,
  \label{eq:shared_moe}
\end{equation}

where $\mathbf{s}_i = G_i(\mathbf{h}_i)$ denotes the routing scores. 
As illustrated in \autoref{fig:moe_arch}(b), cross-layer expert sharing reduces the number of unique parameters by roughly a factor of~$n$ while keeping the activated parameters unchanged.

\subsection{Pre-gated Routing}

To support deployment on devices with limited memory, \modelname~adopts Pre-gated Routing~\cite{hwang2024pre}.
By shifting the gating computation to the preceding layer, the model can load the parameters of the selected experts in advance, which lowers the memory cost of sparse activation.
As shown in \autoref{fig:moe_arch}(c), the router in the preceding layer, $G_{i-1}$, generates the routing decision for layer $i$.

\begin{equation}
  \mathbf{h}_i' \;=\;
  \sum_{j=1}^{M}
  \bigl[\operatorname{top}\text{-}k(G_{i-1}(\mathbf{h}_i)\bigr]_j\,
  E_{g,j}\!\bigl(\mathbf{h}_i\bigr),
  \qquad
  g = \bigl\lfloor i / n \bigr\rfloor,
  \label{eq:megrez_moe}
\end{equation}

Combining pre-gated routing with cross-layer expert sharing yields two practical advantages on device native deployment.
First, when the router predicts the experts required by the next group of layers that do not share parameters, the design behaves like a standard pre-gated MoE, so the computation and weights loading can be overlapped and pipelined to reduce latency.
Second, within a group whose layers share a common expert pool, an expert that has already been selected need not be re-loaded, allowing the cache size and replacement policy to be tuned to the available hardware.
Together, these properties make it feasible to serve large MoE models in memory-constrained environments.

\subsection{Architecture}

In addition to the aforementioned two improvements, \modelnamepreview~adopts a dense-layer-first architecture similar to DeepSeekV2~\cite{liu2024deepseek}. 
The model consists of 31 layers in total, with the first layer being a dense layer. 
Every three layers form a group that shares the expert parameters of the MoE module, which includes 64 experts per group. 
Among these, the top-6 experts are selected via routing. Additionally, each layer incorporates 4 shared experts. 
The hidden size of the dense layers is set to 10,944, and each expert has a hidden dimension of 1,408.
The model employs the tokenizer of Megrez-3B series~\cite{li2025megrez}.

\section{Training Methodology}
In this section, we present the training methodology of \modelnamepreview, which is pre-trained on a diverse dataset containing 5 trillion tokens from multiple domains. Following pre-training, we perform supervised fine-tuning on millions of samples. Finally, we apply reinforcement learning with verifiable rewards to enhance the model’s reasoning capabilities.

\subsection{Pre-training}

\paragraph{Pre-training dataset curation.}
Our dataset curation process follows a similar approach to the Megrez-3B series~\cite{li2025megrez}, but with significant expansions in both scale and diversity. The dataset comprises 5 trillion tokens drawn from a wide range of domains, including web text, cleaned GitHub code, STEM (Science, Technology, Engineering, and Mathematics) content, books, and synthetic reasoning data.

\paragraph{Multi-stage training.}
We adopt a three-stage training paradigm, progressively increasing the diversity and complexity of the data to enhance \modelnamepreview's performance across various tasks.

\dotitem{
\textbf{Foundational Stage.}
This initial stage focuses on building a robust foundation for general language modeling. The model is trained on 1.5 trillion tokens, using a sequence length of 4,096 tokens. The training data in this stage consists primarily of publicly available web text and code.
}

\dotitem{
\textbf{Knowledge \& Reasoning Augmentation Stage.}
To enrich the model’s knowledge and reasoning abilities, this stage introduces more refined and diverse data sources, including high-quality web text, cleaned code, and reasoning tasks with detailed solution processes. We also incorporate proprietary book datasets, specialized mathematical content, and distilled long-context data. A substantial portion of this stage's data consists of synthesized knowledge and reasoning prompts. The model is further pre-trained on approximately 3 trillion tokens, maintaining the 4K sequence length. The learning rate is reduced by an order of magnitude compared to the previous stage, and a cosine annealing schedule is employed to enable fine-grained learning from premium sources.
}

\dotitem{
\textbf{Long-Context Extension Stage.}
The final stage focuses on extending the model's context length to 32K tokens. Following the approach of the Megrez-3B series~\cite{li2025megrez}, we increase the base frequency of RoPE~\cite{su2024roformer} to 1,000,000. Most of the training corpus in this stage is curated by concatenating samples with distributions similar to those in the previous stages. In addition, we introduce specialized long-context datasets to facilitate effective learning over extended contexts. This stage involves more than 600 billion tokens.
}

\subsection{Post-training}

\subsubsection{Supervised Fine-Tuning}

The dataset used in the Supervised Fine-Tuning (SFT) stage consists of millions of high-quality samples, meticulously curated to enhance both general conversational capabilities and specialized skills across diverse domains, such as mathematics, code generation, tabular data processing, and information extraction. Additionally, this dataset extensively incorporates distilled data and synthesized reasoning samples to refine and optimize data distributions, thereby improving model performance on common tasks and achieving closer alignment with human preferences and task-specific requirements.

The \modelnamepreview~model is fine-tuned for 2 epochs on the SFT dataset using a cosine learning rate scheduler, with an initial learning rate one order of magnitude lower than that used during the pre-training stage. Except for the learning rate, most hyperparameters remain consistent with those of the pre-training phase. To enhance training efficiency and model performance, particularly in handling longer multi-turn conversations, training samples are concatenated into fixed-length sequences of either 4K or 32K tokens. 
Crucially, attention masks are configured to ensure mutual invisibility between distinct samples within concatenated sequences, and positional encodings are likewise reset to prevent information leakage.
This strategy enables efficient batching while maintaining the integrity of individual training instances.

Furthermore, we introduce a novel \textit{turn-level loss} mechanism, analogous to the per-sample loss, specifically tailored for multi-turn dialogue data. In this approach, the loss for each conversational turn is computed independently and normalized by the number of tokens in that turn, rather than by the total sequence length. This \textit{turn-level loss} strategy significantly improves model accuracy, especially in complex multi-turn conversational settings.

\subsubsection{Reinforcement Learning}

We employ \textbf{Reinforcement Learning with Verified Reward (RLVR)} to enhance \modelnamepreview’s reasoning abilities. Our approach begins by constructing a large-scale, open-source dataset comprising math and reasoning tasks, which undergoes rigorous filtering for both quality and difficulty. In particular, we discard questions whose answers cannot be reliably verified through string matching, and we convert multiple-choice questions into open-ended formats to minimize the likelihood of correct guesses.

To enrich data diversity, we employ a model-based synthesis pipeline to generate additional samples, while excluding simpler questions from the prompt set. To ensure a balanced difficulty distribution, we further adjust the dataset such that approximately 50\% of the examples are questions the model fails to answer correctly in all attempts. The final dataset contains roughly 60,000 samples.

For reinforcement learning, we utilize a modified version of the Generalized Reinforcement Learning with Policy Optimization (GRPO) algorithm~\cite{shao2024deepseekmath}, replacing the original GRPO loss with a Proximal Policy Optimization (PPO) loss. We adopt Generalized Advantage Estimation (GAE)~\cite{schulman2015high} with both $\lambda$ and $\gamma$ set to $1$. The reward function is binary, assigning a positive reward for correct answers and a negative reward for incorrect ones. To encourage exploration during training, we set the model temperature to 1.0 in rollout stage and disable KL regularization.

\section{Evaluation}
We evaluate \modelnamepreview~across a diverse range of domains, including general language comprehension, instruction-following, mathematical reasoning, and coding tasks. For each benchmark, we compare \modelnamepreview~against strong baselines with either a comparable number of activation parameters or similar total parameter counts. These include several open-source dense models, Qwen2.5-3B, Qwen2.5-7B~\cite{Yang2024Qwen25TR}, Qwen3-4B, Qwen3-8B~\cite{Yang2025Qwen3TR}, Gemma-3-4B~\cite{Kamath2025Gemma3T}, as well as a proprietary model, GPT-4o-mini~\cite{openai-gpt4o-mini} (version dated 2024-07-18). Detailed evaluation results are presented in \autoref{tab:posttrain-comparison-split}.

\begin{table}[!ht]
    \centering
    \caption{Comparison among \modelname~and other representative models.}
    \label{tab:posttrain-comparison-split}
    \resizebox{\textwidth}{!}{%
        \begin{tabular}{ll c c c c c c c c}
            \toprule
                                                       &           & \textbf{\modelname} & \textbf{Qwen2.5-3B} & \textbf{Qwen2.5-7B} & \textbf{Qwen3-4B} & \textbf{Qwen3-8B} & \textbf{Phi-4-mini} & \textbf{Gemma-3-4B} & \textbf{GPT-4o-mini} \\
            \midrule
            \multicolumn{2}{l}{\# Activated Params (B)} & \textbf{3.0}       & 3.1                 & 7.6                 & 4.0                 & 8.2               & 3.8               & 4.3                 & -                                          \\
            \multicolumn{2}{l}{\# Stored Params (B)}    & 7.5       & 3.1                 & 7.6                 & 4.0                 & 8.2               & 3.8               & 4.3                 & -                                          \\
            \midrule
            \multirow{2}{*}{\textit{General Tasks}}
                                                       & C-EVAL    & \textbf{91.7}       & 68.2                & 76.2                & 72.2              & 77.9              & 40.0                & -                   & 66.3                 \\
                                                       & MMLU-Pro  & \textbf{67.6}       & 43.7                & 56.3                & -                 & -                 & 52.8                & 43.6                & -                    \\
            \midrule
            \multirow{1}{*}{\textit{Instruction Tasks}}
                                                       & IFEval    & 80.2                & 58.2                & 71.2                & 81.2              & 83.0              & 68.6                & \textbf{90.2}       & 80.4                 \\
            \midrule
            \multirow{2}{*}{\textit{Math Tasks}}
                                                       & MATH-500  & 81.6                & 65.9                & 75.5                & 84.8              & \textbf{87.4}     & 64.0                & 75.6                & 78.2                 \\
                                                       & GSM8K     & 83.6                & 86.7                & 91.6                & -                 & \textbf{93.2}     & 88.6                & 89.2                & -                    \\
            \midrule
            \multirow{2}{*}{\textit{Coding Tasks}}
                                                       & HumanEval & 74.4                & 74.4                & 84.8                & -                 & 85.9              & 74.4                & 71.3                & \textbf{87.2}        \\
                                                       & MBPP      & \textbf{88.0}       & 72.7                & 79.2                & -                 & 77.0              & 65.3                & 63.2                & -                    \\
            \bottomrule
        \end{tabular}}
\end{table}

Overall, \modelname~exhibits a notably efficient parameter design. With only 3B activated parameters and a total of 7.5B parameters, it delivers performance that matches or even exceeds that of substantially larger models. For instance, it stores fewer parameters than Qwen2.5-7B and Qwen3-8B, yet outperforms them in several benchmarks. This underscores \modelname’s ability to balance model capacity and computational efficiency, making it particularly well-suited for deployment in resource-constrained environments.

\paragraph{General Tasks.}
To assess general language understanding, we evaluate performance on C-EVAL (ZH)~\cite{huang2024c} and MMLU-Pro~\cite{wang2406mmlu}, using \textit{Exact Match (EM)} as the metric.
Despite using fewer activated parameters, \modelnamepreview~consistently outperforms or matches the performance of significantly larger models in these benchmarks. This reflects its strong general reasoning and language comprehension capabilities, achieved with a lightweight architecture.

\paragraph{Instruction Tasks.}
To evaluate instruction-following ability, we use IF-Eval~\cite{zhou2023instruction} under the \textit{Prompt Strict} setting.
\modelname~achieves highly competitive results, closely tracking or surpassing models with larger sizes. While one model achieves a higher peak score, it does so with reduced consistency across task types. In contrast, \modelnamepreview~strikes a solid balance between model size and instruction-following quality.

\paragraph{Math Tasks.}
For mathematical reasoning, we evaluate the model on MATH-500~\cite{hendrycks2021measuring} and GSM8K~\cite{cobbe2021gsm8k}, using \textit{Exact Match (EM)} as the evaluation metric.
\modelnamepreview~demonstrates strong performance in both datasets, often rivaling or exceeding larger models. Even when slightly outperformed in some cases, it achieves competitive accuracy with far fewer parameters, showcasing its strength in numerical and symbolic reasoning.

\paragraph{Coding Tasks.}
To assess code generation, we use two widely adopted benchmarks—HumanEval~\cite{chen2021codex} and MBPP~\cite{austin2021program}—with \textit{Pass@1} as the evaluation metric.
\modelnamepreview~achieves strong results relative to its size, outperforming all other models in at least one benchmark. These outcomes suggest that the model effectively captures programming logic and structure, making it well-suited for code generation tasks.

Across all evaluated domains, the \modelnamepreview~model delivers consistently strong performance, often surpassing larger models on both general and specialized tasks. These results demonstrate the effectiveness of its architecture and training methodology in achieving high performance with lower computational cost, positioning \modelname~architecture as an efficient and versatile solution for a wide range of applications, particularly in device native scenarios.

\section{Conclusion}
In this technical report, we present \modelname, a novel language model architecture that introduces expert sharing across consecutive layers and pre-gated routing within the Mixture-of-Experts (MoE) framework.
We evaluate \modelnamepreview, the first instantiation of the \modelname~architecture, on a diverse set of benchmarks, including general language understanding, instruction following, mathematical reasoning, and code generation.
Despite activating only 3B parameters per token and having a total of 7.5B parameters, \modelnamepreview~matches or even surpasses the performance of significantly larger models.
It demonstrates strong generalization, solid reasoning abilities, and effective task-specific adaptation, all while requiring only a fraction of the computational cost. These results highlight the effectiveness of our approach, making it a cost-efficient solution for both research and deployment in resource-constrained environments.
Overall, the balanced design of the \modelname~architecture provides a compelling trade-off between model size and performance, making it well suited for device native deployment and a wide range of real-world applications.

\printbibliography

\end{document}